\documentclass{article}

\usepackage[table,dvipsnames]{xcolor}

\usepackage[final]{corl_2026} 

\usepackage{graphics} 
\usepackage{amsmath} 
\usepackage{amssymb}  
\usepackage{soul,color} 
\usepackage{svg}

\usepackage{graphics} 
\usepackage{amsmath} 
\usepackage{amssymb}  
\usepackage{amsfonts}
\usepackage{newtxmath}
\usepackage{mathtools}
\usepackage{graphicx}

\usepackage{tikz}
\usepackage{bm}
\usepackage{tikz-3dplot}
\usepackage{pgfplots}
\usepackage{pgfplotstable}
\usetikzlibrary{arrows.meta,arrows}
\usetikzlibrary{positioning}
\usetikzlibrary{shapes.geometric}
\usetikzlibrary{bending, plotmarks, shapes, pgfplots.statistics, pgfplots.colorbrewer, positioning}
\usetikzlibrary{matrix,calc}
\usepgfplotslibrary{fillbetween}
\pgfplotsset{compat=1.18}

\newcommand{\grayrow}{\rowcolor{gray!10}}

\usepackage{booktabs}
\usepackage{multirow}
\usepackage{tabularx}
\usepackage{makecell}

\usepackage[T1]{fontenc}

\usepackage{balance}
\usepackage{float}
\usepackage{stfloats}
\usepackage{placeins}
\usepackage{wrapfig}

\usepackage{url}

\usepackage{svg}
\usepackage{pifont}
\usepackage{xfrac}
\usepackage{ifthen}

\usepackage{bookmark}
\usepackage{amsmath}
\usepackage{amsfonts}
\usepackage{newtxmath}
\usepackage{amssymb}
\usepackage{graphicx}
\usepackage{float}
\usepackage{svg}
\usepackage[acronym]{glossaries}
\usepackage{url}
\usepackage{blindtext}
\usepackage{anyfontsize}

\usepackage[font=small,labelfont=bf]{caption}
\makeatletter
\renewcommand{\section}{\@startsection{section}{1}{\z@}{-1.4ex \@plus -0.4ex \@minus -0.2ex}{0.8ex \@plus 0.2ex \@minus 0.1ex}{\large\bf\raggedright}}
\renewcommand{\subsection}{\@startsection{subsection}{2}{\z@}{-1.2ex \@plus -0.4ex \@minus -0.2ex}{0.4ex \@plus 0.1ex}{\normalsize\bf\raggedright}}
\renewcommand{\paragraph}{\@startsection{paragraph}{4}{\z@}{0.8ex \@plus 0.3ex \@minus 0.2ex}{-1em}{\normalsize\bf}}
\makeatother
\pgfdeclarelayer{bg}

\title{SNAP: Semantic Navigation Among Pedestrians}

\author{
  Jakob Wolfram$^{1}$, Pascal Roth$^{1,2}$, Fan Yang$^{1}$, Cesar Cadena$^{1}$, Marco Hutter$^{1}$ \\
  $^{1}$Robotic Systems Lab, ETH Zurich \quad $^{2}$NVIDIA \\
  \texttt{jwolfram@ethz.ch}
}

\begin{document}
\maketitle
\vspace{-6mm}
\begin{figure}[H]
    \centering
    \includegraphics[width=\textwidth]{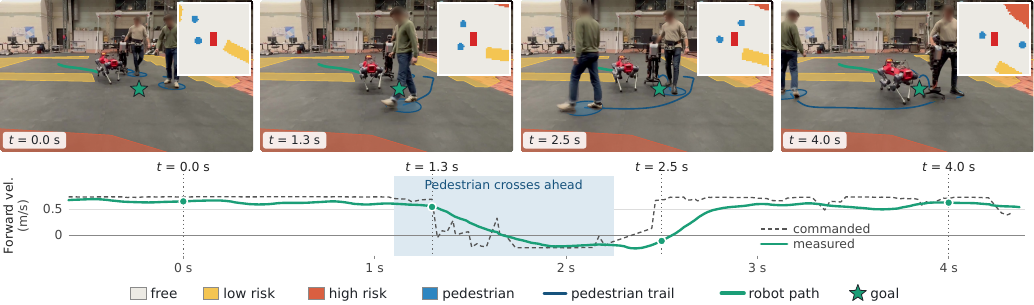}
    \caption{Real-world crowd navigation while following semantic clues with ANYmal-D.
    The policy avoids pedestrians that repeatedly cross the robot's path while also adhering to the underlying semantic structure of the terrain. 
    The robot-centric BEV maps (insets) show how semantics and dynamic obstacles are perceived.
    The commanded forward velocity (bottom) demonstrates the policy slowing down for approaching pedestrians, briefly reversing to avoid a collision.}
    \label{fig:real_world_snapshots}
\end{figure}
\vspace{-6mm}
\begin{abstract}
Navigating crowded urban environments requires robots to avoid pedestrians while adhering to social norms such as remaining on sidewalks. Existing approaches typically address either crowd avoidance or semantic path planning in isolation.
We propose a reinforcement learning (RL) navigation policy that jointly reasons about semantic terrain and dynamic obstacles from a single semantically annotated bird's-eye-view (BEV) map, without relying on explicit human-state estimation.
Our policy outperforms established crowd-navigation and semantic-aware baselines and transfers zero-shot to a legged robot, demonstrating socially compliant navigation in highly dynamic real-world scenarios.
\\
\textbf{Supplementary material:} \url{https://snap-navigation.github.io}
\end{abstract}
\section{Introduction}
\label{chap:introduction}

Robots operating in shared urban spaces must navigate not only safely but also predictably by adhering to social norms. 
More precisely, they are expected to respect personal space, adapt to surrounding pedestrians, and follow norms such as remaining on sidewalks or using crosswalks, otherwise causing disruptive behavior~\cite{review_social_nav}. 
Such behavior is a prerequisite for applications like autonomous last-mile delivery~\cite{Lee2024LearningRA} and, more broadly, the deployment of robots in urban environments.

Crowd navigation, requiring a robot to reach a goal while avoiding dynamic obstacles, has been studied extensively~\cite{alyassi12social, liu_robot_nav, chen2019crowd}. Classical heuristic-based methods involve environment-specific tuning which limits generalization~\cite{Berg_ORCA, Helbing_SFM}. Learning-based methods achieve higher success rates~\cite{alyassi12social} but typically require explicit human-state information such as positions and velocities. Estimating these in real crowds is prone to occlusion and tracking failures~\cite{Guan2025MultiobjectTR}, and the resulting modular pipelines add latency and propagate estimation errors.
In parallel, semantic planners incorporate scene semantics into trajectory planning to follow social norms, such as preferring sidewalks or crosswalks~\cite{roth2024viplanner, wellhausen_safe_nav, sorokin_sidewalk}. However, they treat the environment as static or avoid pedestrians with reactive heuristics.

We bridge both lines of work with an RL policy that observes a robot-centric, semantically annotated bird's-eye-view (BEV) map that encodes terrain classes alongside nearby dynamic obstacles. This unified observation lets the policy reason jointly about \textit{where} to move and \textit{how} to react to the crowd, without explicit pedestrian state estimation. We benchmark it against crowd-navigation and semantic-planning baselines, showing that BEV observations without explicit state estimation achieve better performance, and deploy it zero-shot on an ANYmal-D~\cite{HutterANYmal}.

\begin{figure}[t]
    \centering
    \includegraphics[width=\textwidth]{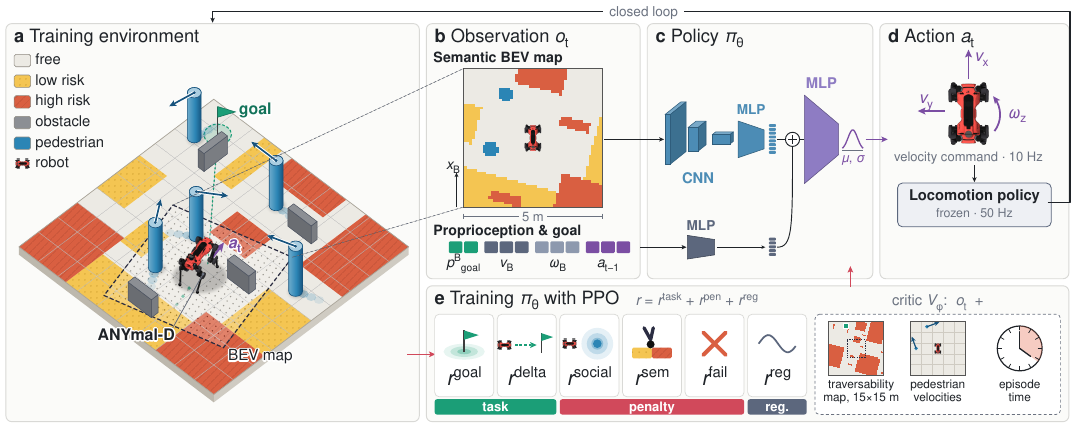}
    \caption{Overview of our framework.
    (a)~Training in procedurally generated environments with semantic terrain of varying risk, static obstacles, and SFM-driven pedestrians.
    (b)~The actor observes a robot-centric semantic BEV map, proprioception, and the target vector.
    (c)~CNN and MLP encoders feed a Gaussian action head.
    (d)~A frozen locomotion policy tracks the generated velocity command.
    (e)~PPO training with the rewards in Table~\ref{tab:rewards} and an asymmetric critic with privileged observations (Table~\ref{tab:observations}).}
    \label{fig:architecture}
\end{figure}

\definecolor{envfree}{HTML}{D8D7D5}
\definecolor{envlow}{HTML}{DBCE71}
\definecolor{envhigh}{HTML}{D69065}
\definecolor{envobstacle}{HTML}{9E9A96}
\definecolor{envped}{HTML}{63A1A8}
\definecolor{envarrow}{HTML}{17537D}
\definecolor{envpath}{HTML}{1B9E77}
\DeclareRobustCommand{\envpatch}[1]{\tikz[baseline=0pt]\draw[black!45, line width=0.3pt, fill=#1] (0,0) rectangle (0.85em,0.65em);}
\DeclareRobustCommand{\envpedestrian}{\tikz[baseline=0pt]\draw[black!45, line width=0.3pt, fill=envped] (0.33em,0.33em) circle (0.33em);}
\DeclareRobustCommand{\envdirection}{\tikz[baseline=0pt]\draw[envarrow, line width=1pt, -{Latex[length=3pt,width=3.5pt]}] (0,0.33em) -- (1.1em,0.33em);}
\DeclareRobustCommand{\envrobotpath}{\tikz[baseline=0pt]\draw[envpath, line width=1.6pt] (0,0.33em) -- (1.1em,0.33em);}

\section{Related Work}
\label{chap:related_work}

Classical methods model pedestrian interactions explicitly, e.g., through attractive and repulsive forces in the Social Force Model (SFM)~\cite{Helbing_SFM}, velocity constraints in Optimal Reciprocal Collision Avoidance (ORCA)~\cite{Berg_ORCA}, or geometric planning~\cite{dynamic_channel}. 
Their hand-crafted rules and assumptions, such as holonomic agents, a shared avoidance policy, or convex shapes, fail to capture the unpredictability of real human motion and limit generalization. 
We therefore learn navigation purely from experience in diverse simulation environments where humans walk based on a randomized SFM formulation.

End-to-end learning methods, mapping sensory inputs to robot actions, can outperform classical approaches~\cite{alyassi12social, Mavrogiannis_core_challenges}. SARL~\cite{chen2019crowd} uses attention-based deep RL for dynamic obstacle avoidance, Liu et al.~\cite{liu_robot_nav} additionally handle static obstacles via occupancy grids and LiDAR angular maps, and DRL-VO~\cite{drlvo} combines a temporal LiDAR buffer with pedestrian velocities.
However, these methods rely on explicit human state estimation, susceptible to occlusion and tracking failures, and treat all free space as equally navigable, ignoring social rules such as sidewalks vs.\ roads. Our policy instead perceives pedestrians in a BEV map, which preserves their spatial arrangement without state tracking, and accounts for terrain semantics.

Semantic approaches restrict robots to safe or socially acceptable terrain, using anomaly detection on multi-modal images to estimate traversability~\cite{wellhausen_safe_nav}, semantic segmentation to separate sidewalks from streets~\cite{sorokin_sidewalk}, or semantic costmaps in trajectory planning~\cite{roth2024viplanner}. These methods largely treat the environment as static and fall back on reactive heuristics when pedestrians block the way. By encoding semantic terrain and dynamic obstacles in a single observation, our policy jointly optimizes both objectives without decoupled planning and avoidance modules.

\section{Method}
\label{chap:method}

\begin{wrapfigure}{r}{0.48\textwidth}
    \vspace{-4pt}
    \centering
    \includegraphics[width=\linewidth]{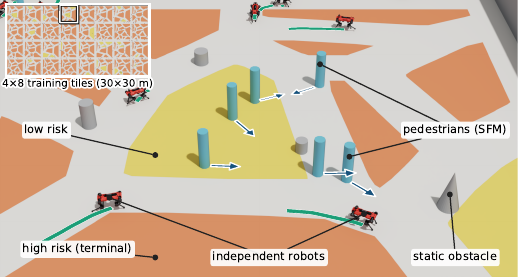}
    \caption{Training environment: $4 \times 8$ tiles of $30 \times 30$~m with free paths \envpatch{envfree}, penalized low-risk \envpatch{envlow} and terminal high-risk \envpatch{envhigh} semantic patches, static obstacles \envpatch{envobstacle}, SFM-driven pedestrians \envpedestrian{} with walking direction \envdirection, and robots sharing the terrain with recent paths \envrobotpath.}
    \label{fig:terrain}
    \vspace{-8pt}
\end{wrapfigure}

Static obstacles of varying shape and pedestrians driven by a randomized SFM~\cite{Helbing_SFM} formulation populate the training environment (Fig.~\ref{fig:architecture}a and Fig.~\ref{fig:terrain}).
To prevent overfitting to deterministic crowd behavior, we randomize the SFM repulsion parameters and assign pedestrians random paths that respect semantic norms (e.g., staying on sidewalks and traversing only low-cost semantic terrain). 
Pedestrians move in groups with probability $p_\text{group}$, and react to the robot by slowing down with probability $p_\text{react}$.

\begin{table}[t]
    \centering
    \begin{minipage}[t]{0.48\textwidth}
        \centering
        \captionof{table}{Asymmetric Observation Space}
        \label{tab:observations}
        \resizebox{\linewidth}{!}{%
\begin{tabular}{@{}llccc@{}}
                \toprule
                \textbf{Modality} & \textbf{Description} & \textbf{Parameter} & \textbf{Actor} & \textbf{Critic} \\ \midrule
                \textit{Proprioception} & Target vector (normalized, body frame) & $x^p_{t,1}$ & \checkmark & \checkmark \\
                 & Base linear/angular velocities & $x^p_{t,2}$ & \checkmark & \checkmark \\
                 & Previously executed action $\mathbf{a}_{t-1}$ & $x^p_{t,3}$ & \checkmark & \checkmark \\
                 & Normalized episode time & $x^p_{t,4}$ & $\times$ & \checkmark \\ \midrule
                \textit{Exteroception} & BEV map ($m \times m$) & $x^e_{t,1}$ & \checkmark & \checkmark \\
                 & Privileged binary traversability map & $x^e_{t,2}$ & $\times$ & \checkmark \\
                 & Occupancy grid with obstacle velocities & $x^e_{t,3}$ & $\times$ & \checkmark \\ \bottomrule
            \end{tabular}%
        }
    \end{minipage}\hfill
    \begin{minipage}[t]{0.50\textwidth}
        \centering
        \captionof{table}{Rewards (S: sparse, D: dense)}
        \label{tab:rewards}
        \resizebox{\linewidth}{!}{%
\begin{tabular}{@{}lll@{}}
\toprule
\textbf{Category} & \textbf{Term} & \textbf{Formulation} \\ \midrule
\textit{Task} & $r_t^\text{goal}$ (S) & $\left(1 + (e_t / \sigma)^2\right)^{-1} \cdot T_\text{rew}^{-1} \cdot \mathbb{I}[t > T_{\max} - T_\text{rew} \lor u_t < p]$ \\
 & $r_t^\text{delta}$ (D) & $d_{t-1} - e_t$ \\ \midrule
\textit{Penalty} & $r_t^\text{social}$ (D) & $-\sum_{i \in \mathcal{H}_t} \exp(-\alpha \| \mathbf{p}_{i,t}^\text{robot} \|_2)$ \\
 & $r_t^\text{sem}$ (D) & $-\sum_{f \in \mathcal{F}} c(\mathbf{p}_{f,t})$ \\
 & $r_t^\text{fail}$ (S) & $-\mathbb{I}[t \ge T_{\max} \lor \mathcal{C} \lor \bigwedge_{f \in \mathcal{F}} \ s(\mathbf{p}_{f,t})]$ \\ \midrule
\textit{Regularization} & $r_t^\text{joint}$ (D) & $-\|\mathbf{\Ddot{q}}_t\|_2$ \\
 & $r_t^\text{lat}$ (D) & $-\min(({}_{B}v_{y,t})^2, 1)$ \\
 & $r_t^\text{act}$ (D) & $-\sum_{i} |a_{i,t} - a_{i,t-1}|$ \\
 & $r_t^\text{mom}$ (D) & $-\sum_{i} |a_{i,t} - (\lambda m_{i,t-1} + (1-\lambda)a_{i,t-1})|$ \\ \bottomrule
\end{tabular}%
        }
    \end{minipage}
\end{table}

\label{subsec:obs_act}
We train the policy using PPO~\cite{ppo} with an asymmetric actor-critic setup~\cite{Pinto2017AsymmetricAC}. Actions $\mathbf{a}_t = ({}_{B}v_x, {}_{B}v_y, {}_{B}\omega)^T \in [-1, 1]^3$ are body-frame velocity commands, scaled and biased toward forward motion.
The actor observes only onboard available observations and no explicit pedestrian states, so it must infer pedestrian configuration and risk from the BEV map. The critic additionally receives privileged information that include pedenstrian states (Table~\ref{tab:observations}, Fig.~\ref{fig:architecture}b-e).

Proprioceptive inputs are encoded by an MLP into a 64-dimensional latent vector, and the BEV map by a three-layer CNN with large kernels and small strides to capture the spatial structure of the static and dynamic obstacles relative to the robot. The concatenated features are mapped to the action distribution by an MLP.

\label{subsec:reward}
Reward shaping is critical to guide the policy toward safe, socially compliant navigation. 
The reward $r_t = r_t^\text{task} + r_t^\text{pen} + r_t^\text{reg}$ is summarized in Table~\ref{tab:rewards}. The task reward $r_t^\text{task}$ combines the time-based terminal reward $r_t^\text{goal}$~\cite{yang2025spatially, he2024agile}, which rewards proximity $e_t$ to the target at the end of the episode and balances goal attainment with exploration, with the dense progress reward $r_t^\text{delta}$, which rewards reductions of the Euclidean goal distance and prevents early convergence to stationary policies. Among the penalties $r_t^\text{pen}$, $r_t^\text{social}$ penalizes proximity to pedestrians $\mathcal{H}$ in the field of view, providing anticipatory guidance beyond binary collisions, $r_t^\text{sem}$ penalizes high-cost semantic terrain under the feet $\mathcal{F}$, and $r_t^\text{fail}$ penalizes collisions and other terminations. Regularization terms ensure smooth, hardware-safe motion~\cite{Rudin2022AdvancedSB, Lee2024LearningRA} by penalizing joint accelerations ($r_t^\text{joint}$) to reduce actuator strain, lateral velocity ($r_t^\text{lat}$) to encourage forward motion, action changes ($r_t^\text{act}$), and deviations from a momentum-filtered action ($r_t^\text{mom}$).

\label{subsec:curriculum}
An episode succeeds if the robot stays within $0.5$\,m of the goal for 2\,s, while it may briefly leave the goal region to evade pedestrians without resetting this timer. 
Episodes fail upon collision, when all feet are on forbidden terrain, or after 60\,s. 
To prevent early training collapse due to overwhelming crowd densities, a curriculum gradually increases the number of pedestrians.

\section{Experiments}
\label{chap:experiments}

\begin{figure}[t]
    \centering
    \includegraphics[width=\textwidth]{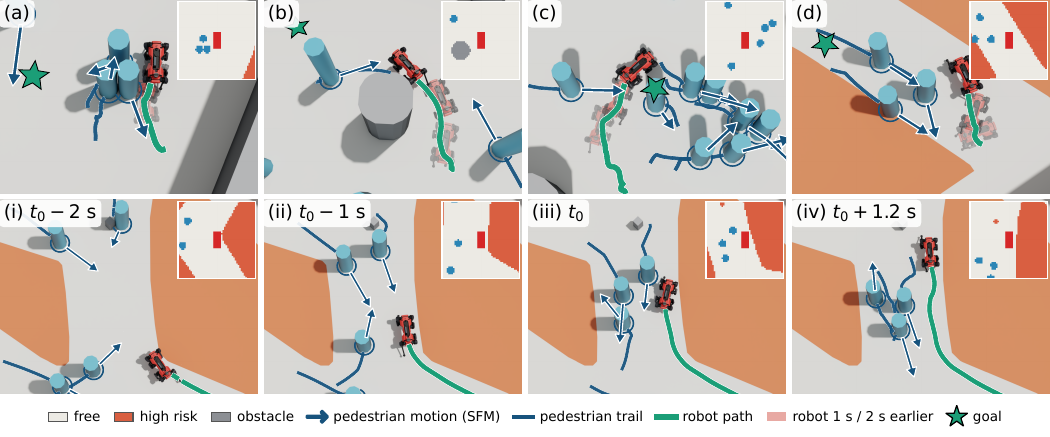}
    \caption{Simulation snapshots. (a)--(d) Avoidance maneuvers in the benchmark settings. Past robot positions demonstrate how the policy navigates around crowds and static obstacles, while adhering to semantically traversability clues.
    (i)--(iv) Temporal sequence showing how the policy passes an oncoming group in a corridor bounded by high-risk terrain. $t_0$ marks the closest approach. The inserts show the $5 \times 5$~m BEV observation of the policy (forward is up).}
    \label{fig:sim_snapshots}
\end{figure}

As, to the best of our knowledge, no method yet addresses semantic crowd navigation, we compare against three baselines, each addressing a subset of the task. 
We evaluate both the original policy and a version retrained and tuned in our setting (marked *). 
Each policy is evaluated for 500 episodes, where an episode succeeds if the robot reaches within $0.5$\,m of the goal (qualitative results in Fig.~\ref{fig:sim_snapshots}). 

We benchmark against SARL~\cite{chen2019crowd} to evaluate our policy's navigation capabilities in purely dynamic environments, as SARL neither supports static obstacles nor semantic terrain. The flat evaluation terrain features five pedestrians and start-goal distances of up to 10\,m.
SARL observes the positions, velocities, and radii of pedestrians. For SARL*, we retrain SARL in our simulator with PPO and continuous actions, as in~\cite{alyassi12social}, and add our regularization and dense guidance rewards (Sec.~\ref{subsec:reward}).
As visible in Tab.~\ref{tab:benchmark}, our policy outperforms both variants at all pedestrian speeds, suggesting that BEV observations are better interpretable than explicit human states for the used omnidirectional robot.

To evaluate the joint capability of static obstacle avoidance and socially compliant navigation, we benchmark our approach against DRL-VO~\cite{drlvo}.
DRL-VO detects obstacles with LiDAR and observes the velocities of pedestrians in its frontal FOV. As it was trained on a TurtleBot2 with a two-dimensional action space, we also retrain it for ANYmal in our setting (DRL-VO*). With start-goal distances of up to 10\,m and pedestrians walking at 0.5--0.8\,m/s, our method outperforms both variants for all crowd sizes (compare Tab.~\ref{tab:benchmark}). We attribute this to the BEV map preserving geometric structure, in contrast to DRL-VO's min- and average-pooled LiDAR features, and to its $360^\circ$ coverage of pedestrians.

\begin{wraptable}{r}{0.40\textwidth}
    \vspace{-10pt}
    \caption{Benchmark Success Rates (SR)}
    \label{tab:benchmark}
    \centering
    \scriptsize
    \setlength{\tabcolsep}{1.5pt}
    \renewcommand{\arraystretch}{0.9}
    \begin{tabular}{cccc}
        \toprule
        \textbf{Speed} $\left[\frac{m}{s}\right]$ & \textbf{SARL} & \textbf{SARL*} & \textbf{Ours}\\
        \midrule
        \grayrow
        0.5 & 0.374 & 0.592 & \textbf{0.932}\\
        0.9 & 0.298 & 0.486 & \textbf{0.776}\\
        \grayrow
        1.3 & 0.190 & 0.384 & \textbf{0.586}\\
        \midrule
        \textbf{Dyn. Obs.} & \textbf{DRL-VO} & \textbf{DRL-VO*} & \textbf{Ours} \\
        \midrule
        \grayrow
        10 & 0.566 & 0.617 & \textbf{0.918}\\
        15 & 0.478 & 0.575 & \textbf{0.854}\\
        \grayrow
        20 & 0.430 & 0.504 & \textbf{0.818}\\
        \midrule
        \textbf{Setting} & \textbf{ViPlanner} & \textbf{Ours*} & \textbf{Ours}\\
        \midrule
        \grayrow
        static & 0.804 & 0.892 & \textbf{0.960}\\
        dynamic & 0.580 & 0.762 & \textbf{0.826}\\
        \bottomrule
    \end{tabular}
    \vspace{-8pt}
\end{wraptable}
To assess performance in semantic terrains, we compare against ViPlanner~\cite{roth2024viplanner}, able to react to dynamic obstacles but originally trained in purely static scenes.
This comparison uses environments that are about 60\% traversable, with start-goal distances of up to 15\,m, consistent with ViPlanner's training range. We evaluate without pedestrians and with 16 pedestrians walking at 0.5--0.8\,m/s. As ViPlanner only uses a forward-facing camera, we also evaluate Ours*, whose BEV map is restricted to a region in front of the robot with a $90^\circ$ FOV.
Both Ours and Ours* outperform ViPlanner in both settings. ViPlanner's success rate drops to 58\% with pedestrians, underscoring the importance of training with moving obstacles.
Finally, the performance difference between Ours* and Ours supports the hypothesis that a $360^\circ$ field of view is essential for navigation in dynamic environments.

We deploy the policy zero-shot at 10\,Hz on ANYmal-D. The robot is localized in a pre-defined semantic map with Open3D SLAM~\cite{jelavic_open3d}, from which the BEV map is extracted. Pedestrians are detected with LiDAR and projected into it. A learned locomotion policy~\cite{miki_nav_policy} tracks the velocity commands.
We evaluate the policy in crowded scenes where pedestrians repeatedly cross the robot's path to disrupt its motion. As shown in Fig.~\ref{fig:real_world_snapshots}, the robot stays within the safe region of the semantic map, adapts its path to pass between crossing pedestrians, and reaches the goal without collision.

\section{Conclusion}
\label{chap:conclusion}
We presented an RL policy for semantically aware crowd navigation that operates on semantically annotated BEV maps instead of explicit pedestrian states, avoiding hazardous terrain while navigating dense crowds. It outperforms specialized crowd-navigation and semantic-planning baselines in simulation and transfers zero-shot to ANYmal-D, running at 10\,Hz.
Future work will extend the $5 \times 5$~m FOV, which limits the reaction time to fast pedestrians, e.g., with a larger LiDAR-based occupancy map, and replace the pre-defined semantic map with onboard semantic BEV reconstruction from cameras to fully bridge into urban space. Code and models will be made publicly available.

%
\acknowledgments{This research was supported by the Swiss National Science Foundation (SNSF) as part of project No.227617 and by grants from NVIDIA (utilized NVIDIA A6000 GPUs). This project has received funding from the European Union’s Horizon Europe Framework Programme under grant agreement No 101070405. 
}
%
\bibliography{iros}
%
\clearpage
\appendix
\section{Ablation Studies}

We examine how the trained policy behaves under increasing difficulty by varying (i)~the number of pedestrians and (ii)~their walking speed.
All experiments take place in a $30 \times 30$\,m environment in which about 60\% of the area is traversable, with start-goal distances of up to 30\,m.
Each configuration is evaluated over 500 episodes with a fixed environment seed to ensure reproducibility, and we report the success rate (SR), i.e., the fraction of successful episodes.
An episode is successful if the robot stays within $\epsilon$\,m of the goal for at least two seconds.
The robot may temporarily leave the goal region to avoid pedestrians without resetting this timer.

Table~\ref{tab:crowd_density} varies the number of pedestrians, whose speeds are sampled uniformly from $[0.5, 0.8]$\,m/s. The SR decreases only moderately as the crowd grows, from 82.0\% with 10 pedestrians to 73.2\% with 16, so the policy copes well with denser crowds.

\begin{table}[ht]
    \caption{Varying Crowd Density}
    \label{tab:crowd_density}
    \begin{center}
        \begin{tabular}{cc}
            \toprule
            \textbf{Num. Dyn. Obs.} & \textbf{SR}\\
            \midrule
            \grayrow
            10 & 0.8200\\
            13 & 0.8040\\
            \grayrow
            16 & 0.7320\\
            \bottomrule
        \end{tabular}
    \end{center}
\end{table}

Table~\ref{tab:crowd_speed} fixes the crowd at 13 pedestrians and varies their walking speed. Here the SR drops sharply, from 84.4\% at 0.5\,m/s to 45.2\% at 1.3\,m/s. Faster pedestrians are therefore much more challenging than larger crowds.

\begin{table}[ht]
    \caption{Varying Crowd Speed}
    \label{tab:crowd_speed}
    \begin{center}
        \begin{tabular}{cc}
            \toprule
            \textbf{Speed} $\left[\frac{m}{s}\right]$ & \textbf{SR}\\
            \midrule
            \grayrow
            0.5 & 0.8443\\
            0.9 & 0.6540\\
            \grayrow
            1.3 & 0.4520\\
            \bottomrule
        \end{tabular}
    \end{center}
\end{table}

We attribute the stronger sensitivity to speed to the robot's limited field of view.
At each step, the policy only observes a $5 \times 5$\,m BEV map centered on the robot, and the robot's own footprint (about $1.5 \times 0.5$\,m) further reduces the free space visible within this window.
A fast pedestrian enters the map and reaches the robot within a very short time, which leaves little room to react.
More pedestrians, in contrast, mainly require more frequent but still timely avoidance maneuvers.

\end{document}